\documentclass[10pt,twocolumn,letterpaper]{article}

\usepackage{wacv}
\usepackage{times}
\usepackage{epsfig}
\usepackage{graphicx}
\usepackage{amsmath}
\usepackage{amssymb}
\usepackage{booktabs}
\usepackage{caption}
\usepackage{subcaption}
\usepackage{float}
\usepackage[table,xcdraw]{xcolor}

\newcommand{\comment}[1]{}

\wacvfinalcopy

\ifwacvfinal
\usepackage[breaklinks=true,bookmarks=false]{hyperref}
\else
\usepackage[pagebackref=true,breaklinks=true,colorlinks,bookmarks=false]{hyperref}
\fi

\begin{document}

\title{Human Saliency-Driven Patch-based Matching \\for Interpretable Post-mortem Iris Recognition}

\author{Aidan Boyd~~~~~~Daniel Moreira~~~~~~Andrey Kuehlkamp~~~~~~Kevin Bowyer~~~~~~Adam Czajka\\
University of Notre Dame, Notre Dame, IN, USA\\
{\tt\small \{aboyd3, dhenriq1, akuehlka, kwb, aczajka\}@nd.edu} \\\\
}

\newcommand{\teaser}{
{
   \begin{center}
        \vskip5mm
        \centering
            \begin{minipage}{\textwidth}
                \centering
                \includegraphics[width=\textwidth]{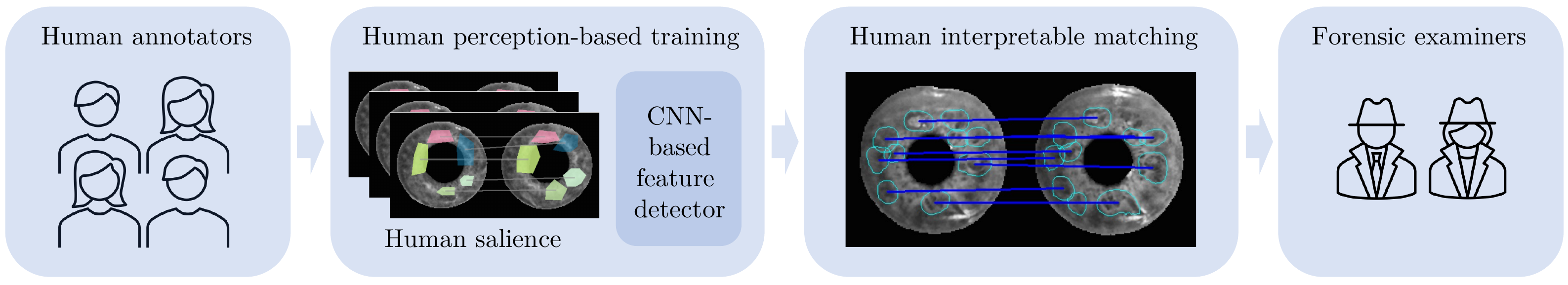}
            \end{minipage}%
            \captionof{figure}{
            Our proposed human-interpretable forensic iris recognition benefits from using a saliency-driven feature detector trained with regions annotated by examiners solving previous forensic iris comparison tasks. Automatically-detected patches are matched and both the overall comparison score, as well as matching feature pairs, are presented to a human examiner.
            }
            \label{fig:teaser}
    \end{center}
}
}

\maketitle
\thispagestyle{empty}

\begin{abstract}
   Forensic iris recognition, as opposed to live iris recognition, is an emerging research area that leverages the discriminative power of iris biometrics to aid human examiners in their efforts to identify deceased persons.
   As a machine learning-based technique in a predominantly human-controlled task, forensic recognition serves as ``back-up'' to human expertise in the task of post-mortem identification. As such, the machine learning model must be (a) interpretable, and (b) post-mortem-specific, to account for changes in decaying eye tissue. 
   In this work, we propose a method that satisfies both requirements, and that approaches the creation of a post-mortem-specific feature extractor in a novel way employing human perception.
   We first train a deep learning-based feature detector on post-mortem iris images, using annotations of image regions highlighted by humans as salient for their decision making.
   In effect, the method learns interpretable features directly from humans, rather than purely data-driven features.
   Second, regional iris codes (again, with human-driven filtering kernels) are used to pair detected iris patches, which are translated into pairwise, patch-based comparison scores.
   In this way, our method presents human examiners with human-understandable visual cues in order to justify the identification decision and corresponding confidence score.
   When tested on a dataset of post-mortem iris images collected from 259 deceased subjects, the proposed method places among the three best iris matchers, demonstrating better results than the commercial (non-human-interpretable) VeriEye approach. 
   We propose a unique post-mortem iris recognition method trained with human saliency to give fully-interpretable comparison outcomes for use in the context of forensic examination, achieving state-of-the-art recognition performance. 
\end{abstract}

\section{Introduction}

The high entropy of iris texture patterns has allowed this modality to rank among the most reliable means of biometric identification of living individuals.  Large-scale iris biometric applications include the national person identification and border
security program Aadhaar in India (with over 1.2 billion pairs of irises enrolled) \cite{Aadhaar}, the Homeland Advanced Recognition Technology (HART) in the US (up to 500 million identities) \cite{USHART}, and the NEXUS system, designed to speed up border crossings for pre-approved travelers between Canada and the US \cite{NEXUS_URL}.  

It was recently demonstrated that matching of live irises with their post-mortem counterparts is feasible \cite{Sansola_MastersThesis_2015,trokielewicz_icb_2016,Trokielewicz_TIFS_2019,Bolme_BTAS_2016}, and -- depending on environmental factors -- can be viable even several weeks after death. This discovery
opened new research areas in forensic iris recognition, with applications that can have huge impact in the context of accidents, natural disasters and combat zones. While automatic image processing and matching tools frequently support forensic examiners, the final decision is made by a human expert. This is why human interpretability of the algorithm's decisions is essential. 

This paper proposes a human-interpretable iris recognition method designed specifically for post-mortem (forensic) iris recognition. The core novel component is the feature detection model trained using image regions annotated by human examiners as salient to their decision-making. 
Instead of narrowing down the interpretable feature selections to several known anatomical iris features (such as collarette, Fuchs' crypts, contraction folds, or Kruckman-Wolfflin bodies), we asked 283 humans to compare iris image pairs and to annotate {\it any} visual features within the iris annulus that support their decision on whether the image pair is a match. The deep learning-model was trained with human-annotated patches to locate iris image regions similar to those selected by humans. Those automatically-detected local regions are then matched using a modified (regional) version of the human-driven binary statistical image features descriptor \cite{Czajka_WACV2019}. Figure \ref{fig:teaser} illustrates the proposed method's pipeline.

This approach may seem to be similar to a traditional keypoint-based iris matching, \eg employing SIFT or SURF descriptors \cite{Alonso-Fernandez_BIDS_2009,Ignat_PCS_2020}, but the {\bf fundamental difference from keypoint-based approaches} lies in the human-driven feature detector, which has several advantages. First, it detects features that are closer to those which human experts would choose. Since these features are more task-driven, they may have better discrimination power and thus a lower number of features is necessary for a high-confidence matching (on the order of a dozen) compared to a general-purpose keypoint-based solutions (which needs hundreds of features to offer high-confidence matching, as shown through experimentation). This makes the visualization less cluttered and more human-interpretable. Second, the proposed method detects regions that have specific shapes, and not just their central locations, which also aids human interpretability.
Last but not least, the matching performance of the proposed method obtained on a sequestered dataset of post-mortem iris images compares very favorably with state-of-the-art methods: it is slightly worse than one of the non-human-interpretable methods \cite{Czajka_WACV2019}, but beats the commercial VeriEye matcher \cite{Verieye} as well as all human-interpretable iris recognition methods known to us, and included in these evaluations.

In summary, the {\bf main contributions} of this work are:
\begin{enumerate}
    \item[(a)] A novel human-driven, human-interpretable iris regional-based matching method, designed specifically for forensic applications;
    \item[(b)] A database of human-annotated iris features, along with the human classification decisions for comparing pairs of post-mortem iris samples; in addition to this paper's reproducibility purposes, this data can serve as a useful resource for studying human-machine pairing in the context of forensic iris recognition;
    \item[(c)] Trained models and source codes of the proposed method, able to be applied in both forensic and live human-interpretable iris recognition.
\end{enumerate}

\section{Related Work}

\paragraph{Forensic Iris Recognition.} Forensic iris recognition was long believed to be impossible, due to incorrect assumptions about the pupil dilation after death, the cornea becoming cloudy \cite{Daugman_BBC_2001}, or even the entire iris decaying only minutes after death \cite{Szczepanski_CISIM_2014}. These assumptions were debunked by Sansola \cite{Sansola_MastersThesis_2015}, who demonstrated that perimortem (image acquired just before death) to postmortem iris matching is possible, and who observed correct matching results for at least 70\% of cases when only postmortem images were compared (depending on time after death). Other groups confirmed the feasibility of forensic iris recognition, with time period after demise ranging from a few days \cite{Bolme_BTAS_2016} (outdoor conditions during summer) to several weeks \cite{Trokielewicz_TIFS_2019,Sauerwein_JFO_2017} (mortuary or winter-time outdoor conditions). 
Several post-mortem iris recognition datasets are available to researchers, created by Trokielewicz \etal \cite{trokielewicz_icb_2016,Trokielewicz_TIFS_2019}. These datasets are accompanied by emerging, but non-human-interpretable, post-mortem iris recognition methods, following the well-known iris code approach (with domain-specific filtering kernels) \cite{trokielewicz_ivc_2020,troki_wacv2020}. The existence of a survey devoted to post-mortem iris recognition \cite{Boyd2020} suggests that this research area has gained momentum, and results may contribute to large-scale forensic applications such as the FBI's Next Generation Identification (NGI) service \cite{FBI_NGI}.

\paragraph{Explainable and Region-Based Approaches.}
Iris recognition results have historically been opaque to human interpretability.
Daugman's \cite{Daugman_TIFS_2016} mathematically-elegant explanation of iris recognition's high discriminative power does not lead to intepretability in the context of intuitive features easily recognized by humans. 
This has led to work on interpretable approaches that point a human examiner to elements of the matching results that aid in explaining their decision of whether two images are from the same iris. 

Active application of deep learning methods to iris recognition changed this situation rather marginally. Among various Convolutional Neural Network-based approaches known to us \cite{Liu2016DeepIrisLP,Gangwar2016DeepIrisNetDI,Minaee2016AnES,Nguyen2017,Zhao_2017_ICCV,Ahmad2019,Wang2019IrisIS,Yang_2021_WACV,boyd_triplet}, only one proposed a human-interpretable output.
This is in a form of Class Activation Map overlaid on post-mortem iris images to suggest to the human examiner regions that were salient to the model \cite{Kuehlkamp_WACVW_2022}. There have been previous attempts that approach iris recognition via keypoint-based matching,
which can be more interpretable than iris codes. Important works include using Scale-Invariant Feature Transform (SIFT) for iris image retrieval \cite{Sunder_ICPR_2010}, combining Speeded-Up Robust Features (SURF) with wavelet-based texture descriptors \cite{Ignat_PCS_2020}, and leveraging anatomical properties of the iris, such as crypts and anti-crypts \cite{Shen_WACV_2014}. None of the previous methods, however, were designed specifically for forensic applications.

{\bf Our work is different from past efforts} in the following respects.
One, our technique is designed specifically for post-mortem iris recognition, using a large dataset of images acquired from 430 deceased subjects.
Two, our feature detector is trained using features annotated by humans as salient for their decision on matching, and so it is guided toward detecting features salient for humans.
Human-interpretable results are essential if iris matching is to serve in forensic applications.

\section{Features for Human Iris-Match Decisions}
\label{sec:annotation_collection}

To understand what features are useful to human examiners for post-mortem iris matching, we collected annotations from humans performing an iris-matching task\footnote{All data collection was done under an approved IRB protocol that allows for distribution of the data to the research community.}.
Following a process similar to the ACE-V protocol used in fingerprint matching \cite{VanderKolk-2011}, data acquisition took
place in two steps.
The first step is {\it Match Evaluation}, as in the ``evaluation'' step in ACE-V, during which ``the final determination as to whether a finding of individualization, or same source of origin, can be made'' (cf. Sec. 9.3.2 and 9.3.3 in \cite{VanderKolk-2011}). In this step, the collection of annotation data was made from subjects matching iris images in the absence of any prior knowledge about the source of samples. The second step is {\it Match Verification}, as in the ACE-V ``verification'' stage, which is ``independent examination by another qualified examiner resulting in the same conclusion'' (cf. Sec. 9.3.5 in \cite{VanderKolk-2011}). In this step, the annotations collected in the first step were presented to new subjects for them to either agree or disagree with, and to supply annotations supporting their decisions.
The annotation tool for both step 1 and step 2 is shown in the supplementary materials. 

Iris images used in our experiments are a combination of a publicly-available post-mortem dataset \cite{Trokielewicz_TIFS_2019} and a dataset collected in a medical examiner's office, which has been submitted to the National Archive of Criminal Justice Data (NACJD) archives \cite{NACJD} and can be requested from the NACJD for research purposes.
All the iris images used in our work are available to other researchers.

\paragraph{Step 1: Match Evaluation.}

Subjects were presented with a pair of post-mortem iris images, and asked to decide whether the images are from the same eye or different eyes.
Once this decision was made, they were
asked to annotate features salient to their decision.
If the decision was that the images are from the same eye, they were asked
to annotate at least 5 pairs of corresponding features between the images. These will be referred to
as matching features.
If the decision was that the images are from different eyes, they were
asked to highlight at least 5 regions on either iris that are present in one image but not
the other. These are called non-matching features. 
There is also a ``Don’t Know'' option to address inconclusive cases. 
When this option was selected, they could then annotate either non-matching features, matching features or combinations of both. 

The step 1 data was collected from 152 human annotators.
All of the annotators are individuals associated with the
University of Notre Dame. 
Each annotator was presented with 20 image pairs: 10 from the same eye (``genuine'' pairs)
and 10 from different eyes (``impostor'' pairs).
The pairs were presented in an order randomized for each annotator. Within the genuine/impostor categories, sampling was performed
based on the post-mortem interval (PMI, time in hours since death). Pairs were curated such that
at least one of the eyes in the pair is from a low-PMI range to
make sure that the matching of artefacts such as specular highlights and wrinkles is minimized as
these appear less frequently in lower PMI samples. It is also a more likely scenario in practice that lower
PMI images are compared to higher PMI images.

\paragraph{Step 2: Match Verification.}

In a verification trial, a new annotator is presented with an image pair from a previous (matching) trial, the previous annotator's decision for that image pair, and a random subset of the previous annotations.
Some data cleaning was performed prior to the verification trials, to remove incorrectly-completed annotated pairs from the previous matching trial.
These new annotators were asked to make the same decision as in the matching trial: do the two images come from the same eye, or eyes of different persons?
In the same manner as a matching trial, annotators were also required to annotate five feature-match 
pairs/non-matching features. 
The annotator on a verification trial could agree or disagree with the results from the previous matching trial.
The inclusion of the annotations from the previous matching trial should serve to highlight regions that lead to different decisions.

Note that an annotator participating in a sequence of verification trials sees results of previous match trials by different annotators. One research question behind the verification trials is to find out if knowing the results of a previous match trial for a particular pair of images leads to better
annotations and more accurate classification.

The step 2 annotations were collected from 131 new subjects using Mechanical Turk (MTurk). 
Restrictions on the MTurk workers include that the worker (a) be an MTurk ``Master'', meaning they had an exceptional approval rating, and (b) be located in a native-English-speaking country, to reduce communication errors in the instructions or instructional video. The annotations were visually
inspected to remove blatantly erroneous samples, resulting from: tool
malfunction, obvious misunderstanding of the task, and apparent ``speed-runners'', who gave minimal
effort to move through the task as quickly as possible. Of the 2620 pairs shown in the second
round, 89 (just over 3\%) were deemed unusable. For each matching-trial annotation, there is an acceptable-quality verification-trial annotation.

\paragraph{Annotation Results.}

\comment{
\begin{table*}[]
\begin{center}
\begin{tabular}{|c|c|c|}
\hline
                             & \textbf{Incorrect to Correct} & \textbf{Correct to Incorrect} \\ \hline
\textbf{Genuine to Impostor} & 153                           & 46                            \\ \hline
\textbf{Impostor to Genuine} & 92                            & 109                           \\ \hline
\textbf{Unsure to Genuine}   & 27                            & \cellcolor[HTML]{333333}      \\ \hline
\textbf{Unsure to Impostor}  & 63                            & \cellcolor[HTML]{333333}      \\ \hline
\textbf{Genuine to Unsure}   & \cellcolor[HTML]{333333}      & 15                            \\ \hline
\textbf{Impostor to Unsure}  & \cellcolor[HTML]{333333}      & 15                            \\ \hline
\end{tabular}
\end{center}
\caption{PLACEHOLDER}
\label{tab:decision_changes}
\end{table*}
}

\begin{table}[!htb]
\begin{center}
\caption{Accuracy of human subjects comparing and annotating pairs of iris images in two steps of the experiment.}
\label{tab:annot_results}
\begin{tabular}{c|c|c}
& \textbf{Step 1} & \textbf{Step 2} \\ 
& (evaluation) & (verification) \\ 
\hline\hline
Overall & 57.3\%  & 60.9\% \\ \hline
Genuine pairs & 36.3\% & 34.9\% \\ \hline
Impostor pairs & 78.4\% & 86.9\% \\ \hline
Inconclusive & 8.9\% & 4.9\% \\ \hline\hline
Number of annotators  & 152  & 131 \\
\end{tabular}
\end{center}
\end{table}

As shown in Table~\ref{tab:annot_results}, accuracy of the verification trials is higher than that of the matching trials, primarily due to impostor pairs of images being classified with higher accuracy in the verification step. Interestingly, for genuine pairs the accuracy decreased slightly from the step 1 (evaluation) trials to the step 2 (verification) trials.
Also, the number of inconclusive decisions decreased greatly in the verification trials, from 8.9\% to 4.9\%.

The inclusion of the decision and annotations from a previous matching trial allowed annotators in a verification trial to make better-informed decisions. Annotators could either agree or disagree with the previous annotation, but the additional context increased the overall accuracy. Because invalid matching-trial annotations were removed, the annotations shown to verification-trial annotators were good examples of correct experimental procedure. Thus, with this guidance the quality of annotations in verification trials increased, as well as the overall accuracy.

\section{Methodology}

As the PMI of post-mortem samples increases, the iris surface area usable for recognition diminishes. This a result of the emergence of decomposition artifacts such as cloudiness or wrinkles \cite{Boyd2020}.
As well, the circularity of the iris boundaries becomes compromised. Thus, to improve the robustness of post-mortem iris matching, the proposed method does not use the traditional assumption of a circular (or elliptical) iris boundary, or that all the iris that is not-occluded by eyelids contains usable texture.
These are factors that make post-mortem iris matching different from and harder than traditional iris recognition. 

One of our approaches to circumvent post-mortem deformations is -- in addition to forensic iris-specific image segmentation -- to try to detect features that are unaffected by the decomposition process. 
Our method detects small regions of usable iris texture in an image, similarly to what humans would do, and then represents these feature patches as unique feature descriptors. The set of feature descriptions for two iris images is then used for matching.

The proposed solution consists of three components: the feature detector, the feature descriptor
and the matching scheme. The goal of the feature detector is to find usable iris texture regions as explained above. The goal of the feature descriptor is to represent the detected regions of iris texture such that they are easily discernible from each other. Given the set of feature descriptions for two iris images, the matching scheme outputs a score for the degree of similarity of the irises. This proposed method is thus referred to as Patch-Based Matching (PBM).

The over-arching design goal for our method is to be 
visually understandable to human examiners. The network we use for feature detection returns a visual representation of where the features are located, and then both the feature description and matching scheme together can show which features are being matched together. 
At the potential trade-off of slightly lower accuracy, our approach is completely transparent about the regions of the images that support the match / non-match decision and presents these results in a human-interpretable way to examiners.

\subsection{Databases}
\label{subsec:datasets}
The first publicly-available dataset for post-mortem iris recognition is the Warsaw BioBase Postmortem Iris v2.0 (Warsaw v2.0) \cite{Trokielewicz_TIFS_2019}. It  consists of 1,200 near-infrared (NIR) post-mortem images from 37 unique cadavers in a  mortuary environment. The PMI ranges from 5 to 800 hours.

Two additional datasets were used in this work, both acquired in an operational medical examiner's setting. The first of these, \textit{dataset 1}, contains $621$ NIR images from 134 cadavers (254 distinct irises).  Images were acquired in sessions of varying PMI, up to a maximum of 9 sessions and 284 hours after death. The second dataset, \textit{dataset 2}, consists of $5,770$ NIR images from 259 subjects. The longest PMI in this dataset is 1,674 hours (69 days), captured at 53 different PMI sessions. Warsaw v2.0, {\it dataset 1} and {\it dataset 2} are entirely subject-disjoint. Also, while {\it dataset 1} and {\it dataset 2} were collected at the same institution, Warsaw v2.0 was collected at a different institution.
Warsaw v2.0 is combined with {\it dataset 1} to create what is referred to as the ``combined dataset'' used for training and validation, and to collect human decisions and annotations, as described in Sec.~\ref{sec:annotation_collection}. {\it Dataset 2} is held out during training and validation and acts as a subject-disjoint test set. 

\subsection{Data Preprocessing}

\paragraph{Annotation Data Preprocessing.}  
For this training, only annotations from \textbf{correct decisions} about \textbf{genuine pairs} 
are shown to the network. Because each image was annotated multiple times in different pairs, bountiful human-derived ground truth is collected. It was decided that supplying the same image to the network with different annotations from each annotator that saw the image could 
hinder effective learning. Conversely, simply using all feature annotations for an image would be too much redundant information, as many people may have annotated the same feature. Thus, to conserve resources for training and remove redundant annotations, a method of aggregating to one ground truth set of annotations was applied. This was achieved by first collecting all sets of annotations for a given image. Next, we take all overlapping annotations and if there is an overlap of greater than 50\% area, we remove the smaller feature annotation. This leaves us with the minimum feature set where the overlap between any two annotations is no greater than 50\%. 
The resulting set of iris images with associated correct annotations contains 716 images, split in a subject-disjoint manner to end up with 518 images in the train set and 198 images in the validation set (70\%/30\% proportion).

\paragraph{Image Preprocessing.} Using a post-mortem-iris-specific application of SegNet \cite{trokielewicz_ivc_2020}, the iris images are first segmented 
and cropped to $256\times256$ pixels around the detected iris. The segmentation mask is used to set all regions not corresponding to the iris texture to zero (black in the image). 
Contrast-limited adaptive histogram equalization (CLAHE)  is applied to the cropped image to accentuate the iris texture, as reported in \cite{Moreira_WACV2019} to be an effective image enhancement means in case of forensic iris recognition.

\subsection{Feature Detection}

\begin{figure}[t]
        \centering
          \begin{subfigure}[b]{1\columnwidth}
      \begin{subfigure}[b]{0.49\columnwidth}
          \centering
            \includegraphics[width=1\columnwidth]{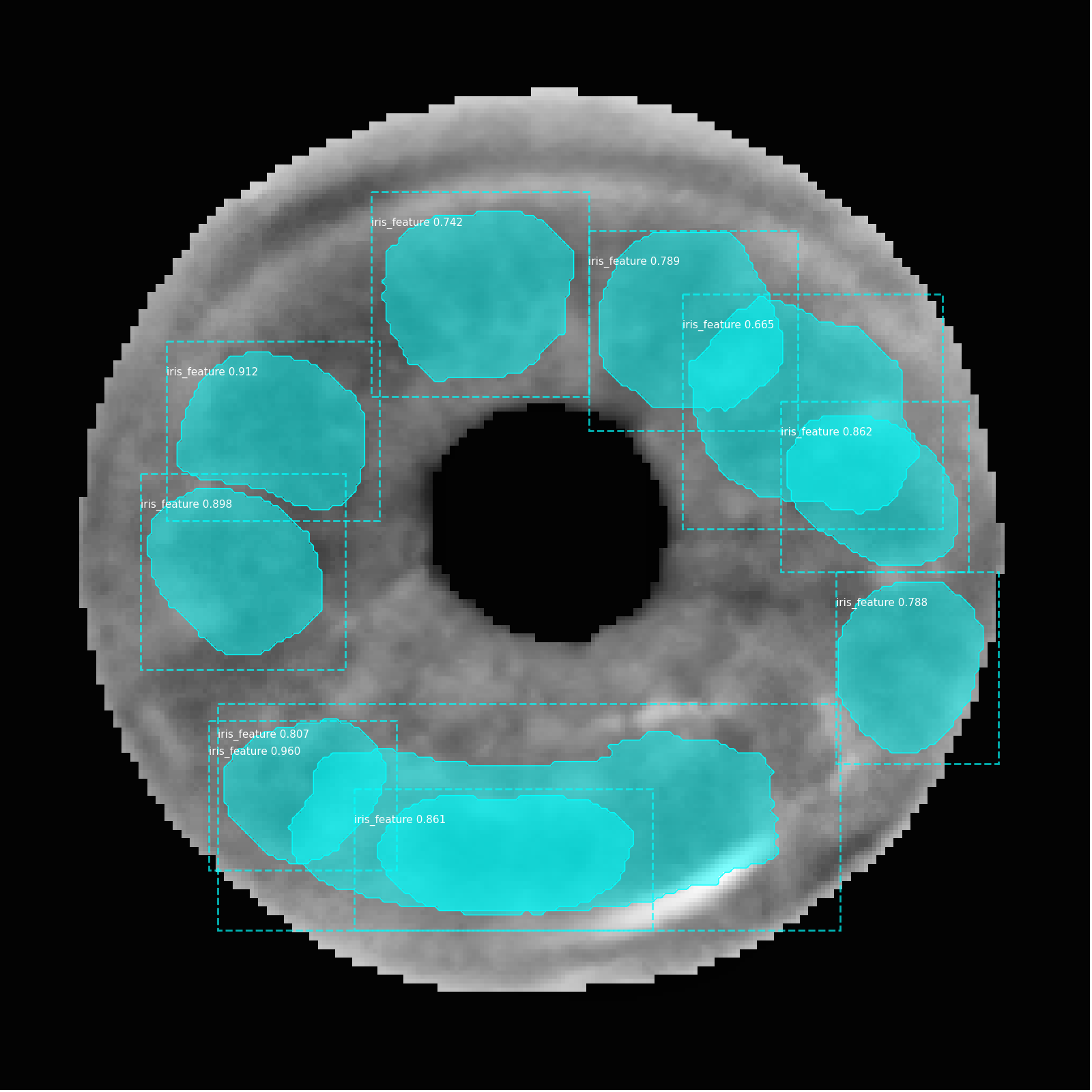}
      \end{subfigure}
      \begin{subfigure}[b]{0.49\columnwidth}
          \centering
          \includegraphics[width=1\columnwidth]{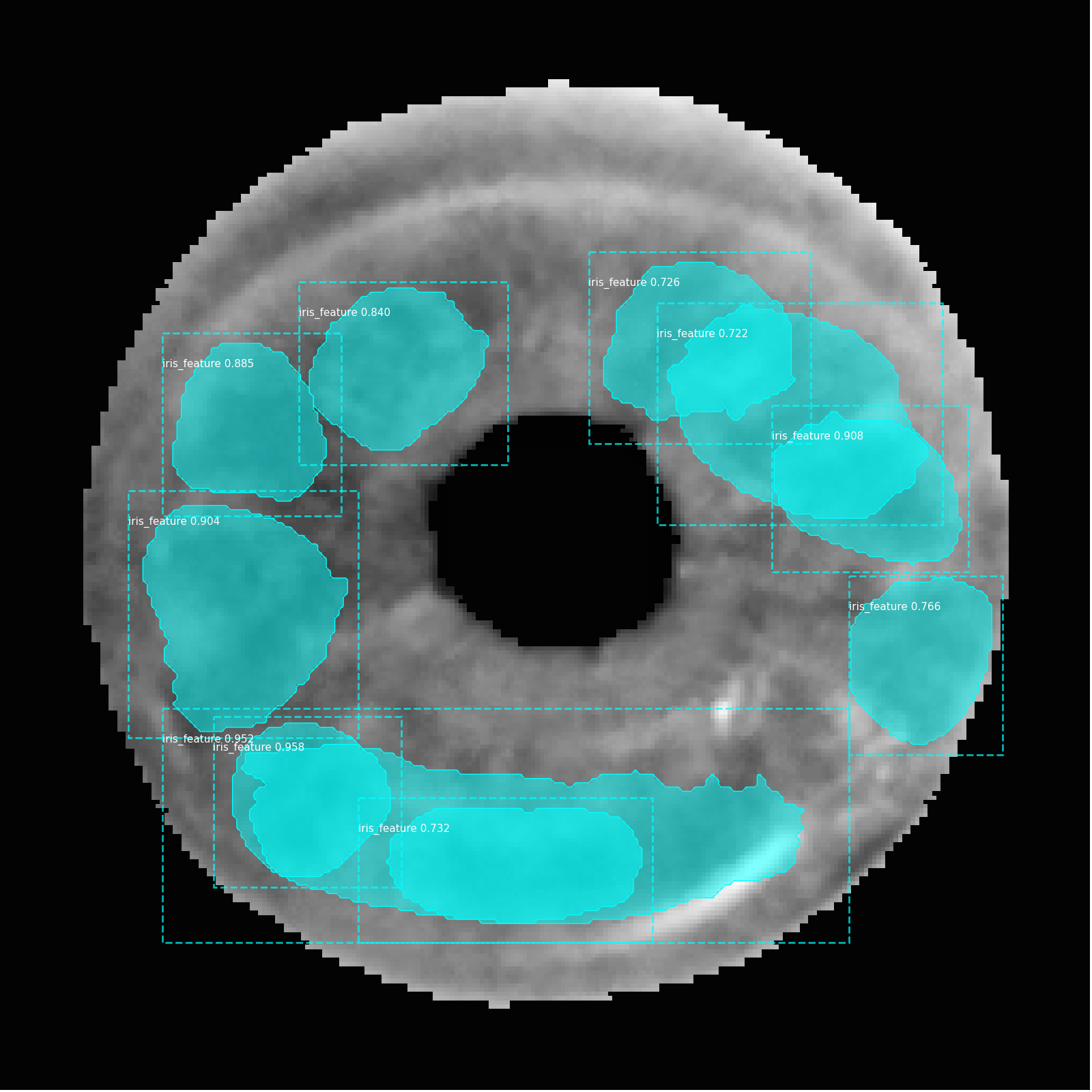}
      \end{subfigure}
      \caption{Genuine Iris Pair}

  \end{subfigure}\vskip3mm
  \begin{subfigure}[b]{1\columnwidth}
      \centering
      \begin{subfigure}[b]{0.49\columnwidth}
          \centering
          \includegraphics[width=1\columnwidth]{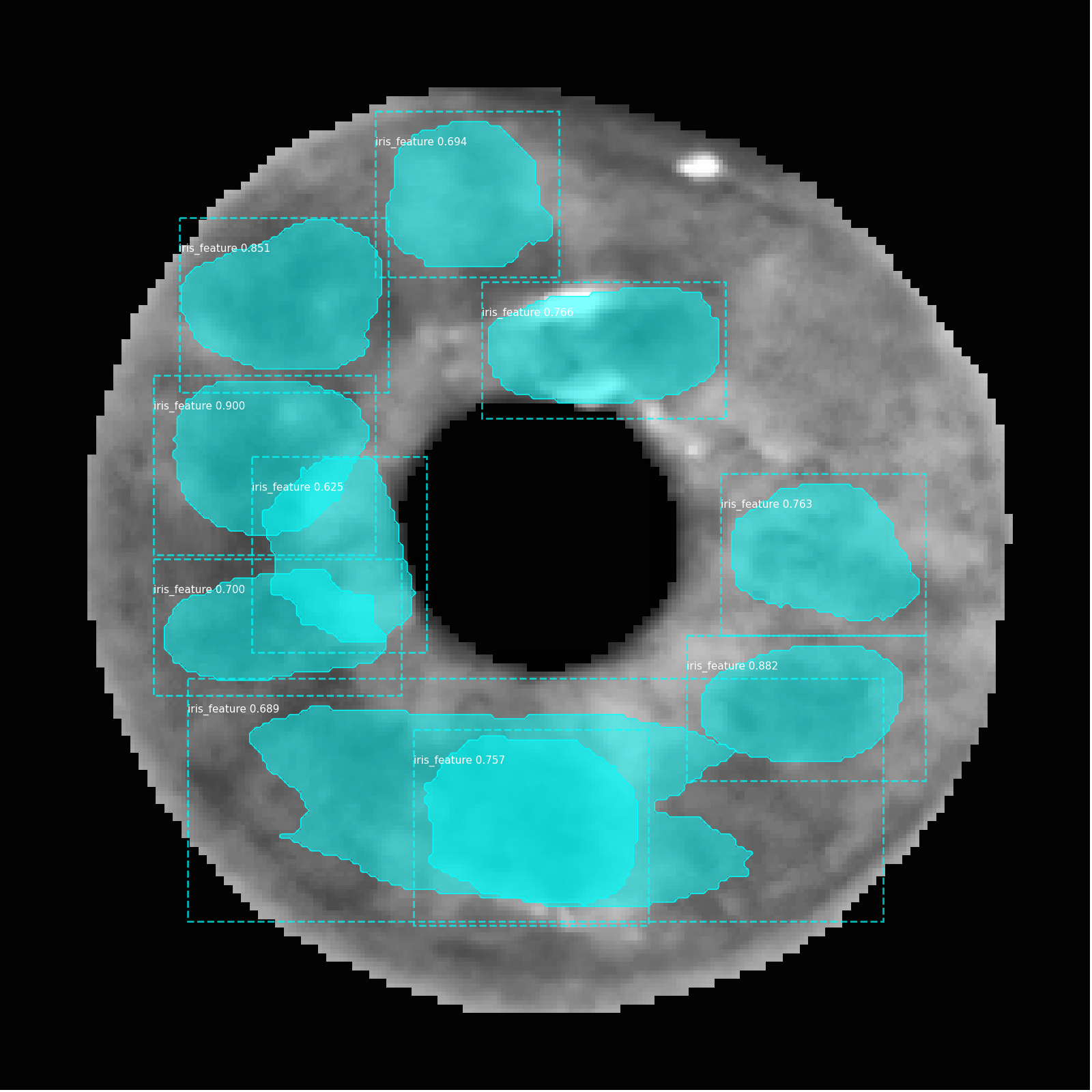}
      \end{subfigure}
     \begin{subfigure}[b]{0.49\columnwidth}
          \centering
          \includegraphics[width=1\columnwidth]{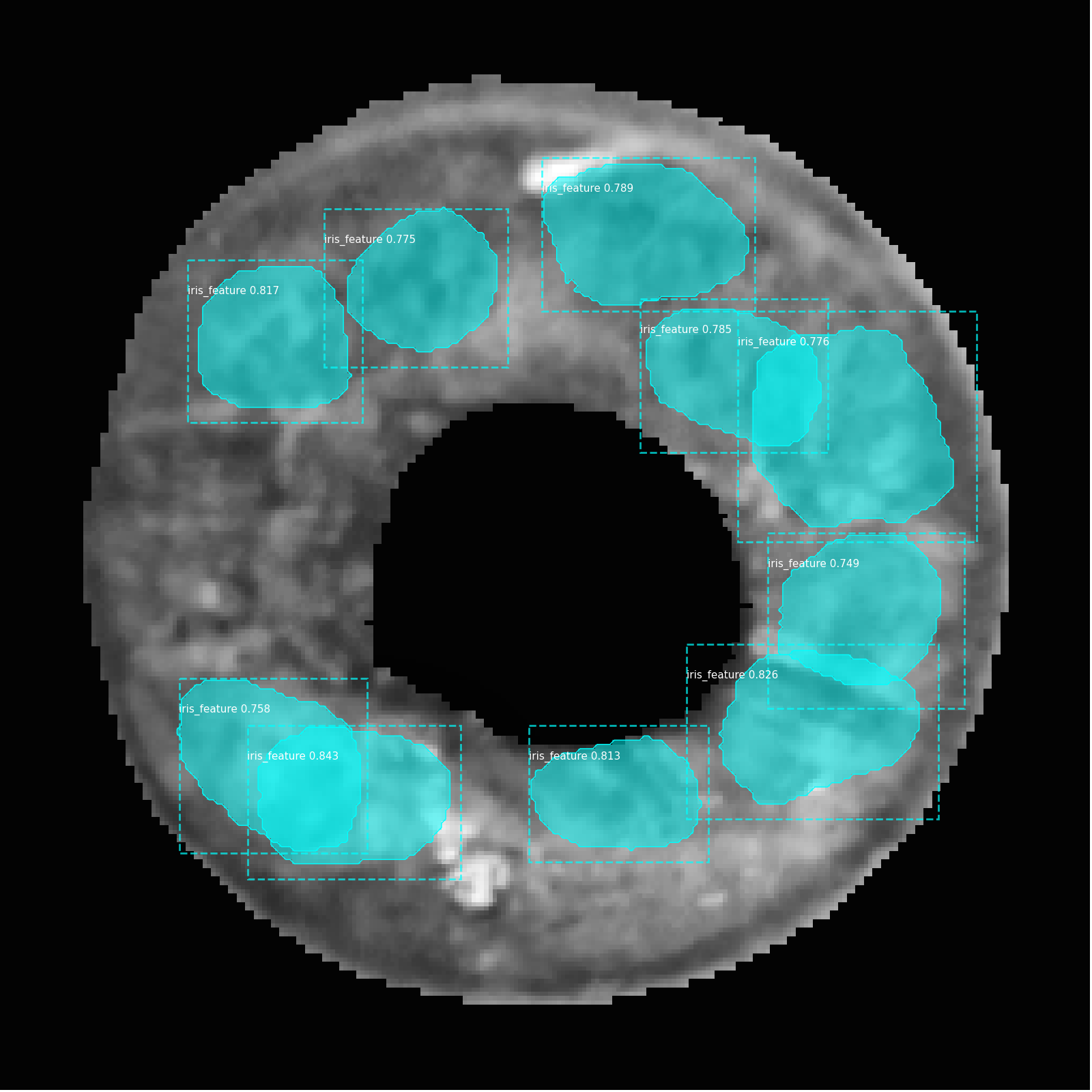}
      \end{subfigure}
      \caption{Impostor Iris Pair}
  \end{subfigure}\vskip-1mm
        \caption{Output of the human-guided MaskRCNN-based feature detector for a genuine pair (a) and an impostor pair (b). Cyan patches are automatically-detected features. }
        \label{fig:detection}
\end{figure}

The MaskRCNN instance segmentation architecture with a ResNet50 \cite{he2016deep} backbone is 
trained to detect individual features present in the iris. 
In addition, a confidence score is also returned
and can be used to rank the detected features. 
Two examples of iris images with detected features can be seen in Fig. \ref{fig:detection}.

A noteworthy point is that the data is annotated in a pair setup, whereas only individual images
are used for the MaskRCNN model training. The rationale is that if pairs are annotated rather than individual images, only features that can be used for matching will be annotated.
If all features were annotated, some might not be useful for recognition. The goal from this network is to determine regions with all good features for matching.

\paragraph{Experimental Parameters.} Due to the limited size of the dataset, extensive augmentation is performed. The augmentations used include left-right flip, up-down flip, $\pm 30$ degrees rotation and Gaussian blurring. This set of augmentations makes sense in the post-mortem iris recognition domain (for instance, up-down flips or severe rotations may happen when a deceased person is approached by an operator from different angles; this is almost not observed in case of live iris recognition, where subjects' eyes are usually positioned horizontally and aligned with the sensor).
All layers in the model were trained for 10 epochs using a learning rate of 0.001. After 10 epochs, the learning rate is divided by 10 and the network head layers are fine-tuned for a further 10 epochs. The optimizer for the network was Stochastic Gradient Descent (SGD). To enable experimental replication, the MaskRCNN specific parameters used to train the model are included in the supplementary materials. The trained model weights are also released with this work.

\subsection{Feature Description}

\begin{figure}
        \centering
        \includegraphics[width=\columnwidth]{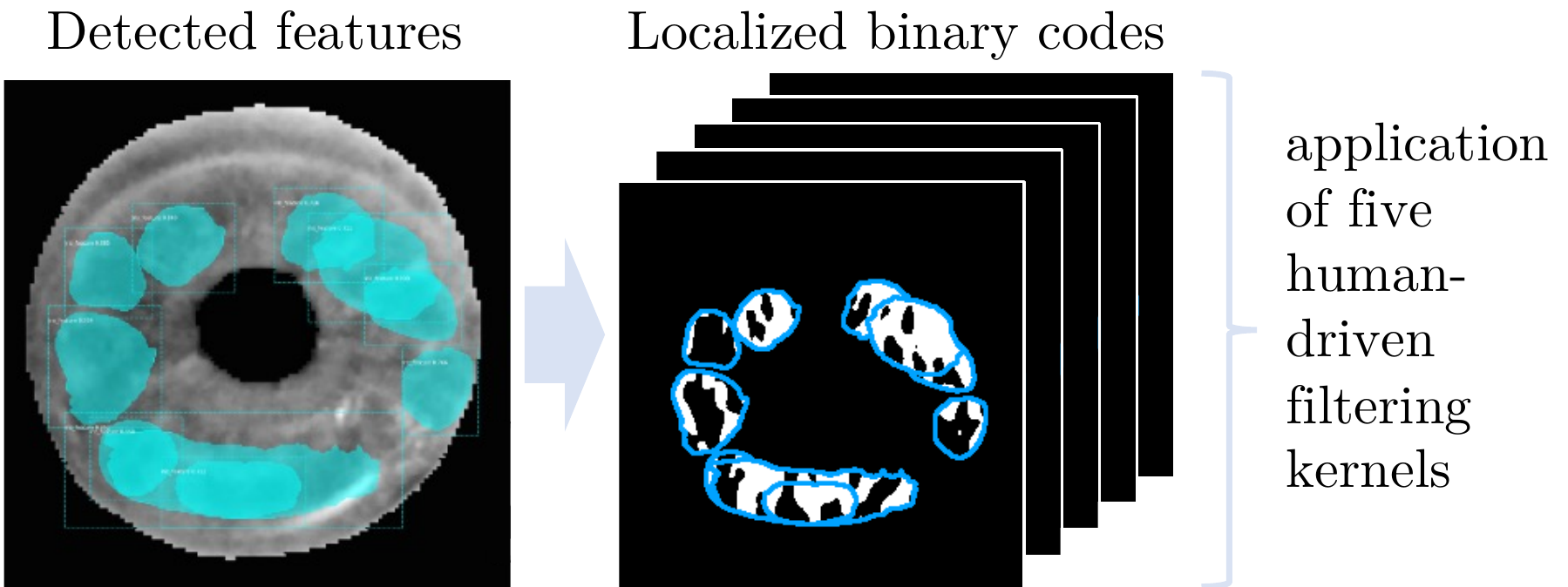}
        \caption{Visualizations of local iris patches encoded with a modified human-driven BSIF method.}
        \label{fig:description}
\end{figure}

Based on the performance of post-mortem matchers in the recent survey by Boyd \etal \cite{Boyd2020}, the best current performing method is using human-driven BSIF \cite{Czajka_WACV2019}. However, this approach performs iris matching in a non-interpretable manner such that it is unclear what features present in both irises lead to a match. 
The proposed feature description for our work aims to leverage the proven performance of the human-driven BSIF method and combine it with our human-interpretable feature detection. In a traditional deep-learning based approach, input images need to be of a specific size, so either features must be extracted of that size only or resized. This ignores both the shape and scale of the feature and can lead to false matching. As the human-driven BSIF approach is not deep learning based, there is no size constraint on features and thus structural integrity of detected features is preserved.

To achieve this integration to our method, the cropped images are encoded in the same BSIF format as was found to be optimal by \cite{Czajka_WACV2019}. That is, we apply five filtering kernels of size $17\times17$ pixels, learned using eye-tracking data. Using the feature set detected with the MaskRCNN model, the feature description is the extracted region of that feature on the BSIF encoded image, as shown in Fig.~\ref{fig:description}.

\subsection{Matching Scheme}

\begin{figure}[t]
                \centering
          \begin{subfigure}[b]{1\columnwidth}
      \includegraphics[width=\columnwidth]{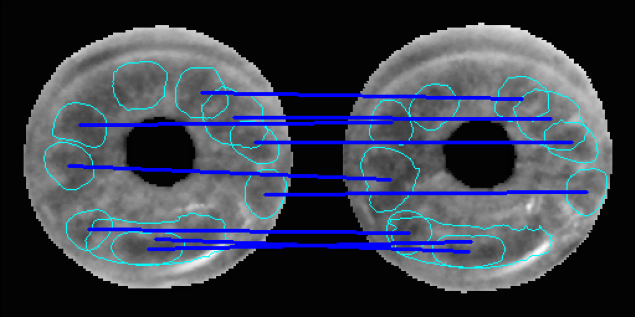}
      \caption{Genuine Iris Pair - Match}

  \end{subfigure}\vskip3mm
  \begin{subfigure}[b]{1\columnwidth}
      \centering
      \includegraphics[width=\columnwidth]{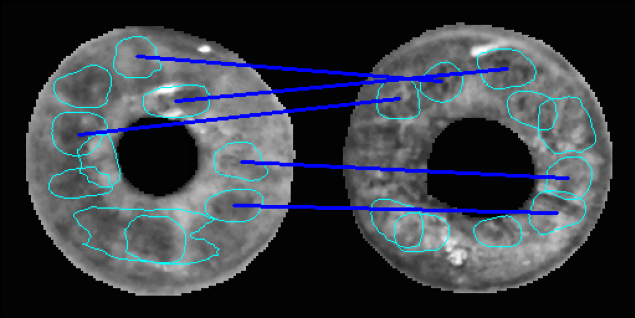}
      \caption{Impostor Iris Pair - Non-match}
  \end{subfigure}\vskip-1mm
        \caption{Output of Patch-Based Matching. Pairs of features that are matched as being the most similar are linked by the dark blue lines. The genuine pair (a) shows parallel lines linking features resulting in a match, whereas the impostor pair (b) has crossing lines and a non-match result. The human examiner can quickly verify the algorithm's result by examining the proposed matching features.}
        \label{fig:matching}
\end{figure}

Given two sets of detected features, the first step is to determine the set of all valid matches. For a pair of features (one on each image) to be considered, they must be within $\pm20$ degrees of rotation of one another. This is determined by establishing the center point of both irises using the segmentation mask. Using the relative position of these iris centers to the center point of the detected features, the angle can be determined. 

The distance metric used to determine the closeness of two features is the Hamming distance (HD) between the two feature descriptions (binary iris codes obtained for each filter). Because five filters are used to get the iris code, the final distance is the mean of the HDs calculated for each of the iris codes. For this distance calculation to work, features need to be the same size. Thus, the largest possible overlap between the two features is calculated and a full iteration of all possible combinations of overlapped features is performed. The smallest distance found in any iteration is accepted as the score for that pair. To insulate against edge cases, the area of the maximum possible overlap must be greater than 50\% the area of the smaller feature.

Once the list of all valid matching features is established, this list is reduced to ensure that each feature can only be used in one match. 
This is done by ranking all valid matches based on their distance apart, smallest to largest. Once a feature has been used in a pair, neither of those features can be used again in other pairings. The initial sorting ensures only the strongest pairings are maintained. The final matching score for the two sets of detected features is the average distance of the five most similar feature pairs, or however many there are if less than five. Thus, the closer the score to zero, the more similar the feature pairs in the images and the more likely it is a genuine match. An example of the output of the tool for both a genuine and impostor pair can be seen in Fig.~\ref{fig:matching}. Matching features are clearly articulated in a human-interpretable manner.

\section{Evaluation}

\begin{figure}[t]
    \begin{center}
       \includegraphics[width=1.0\linewidth,trim={12 0 12 0}]{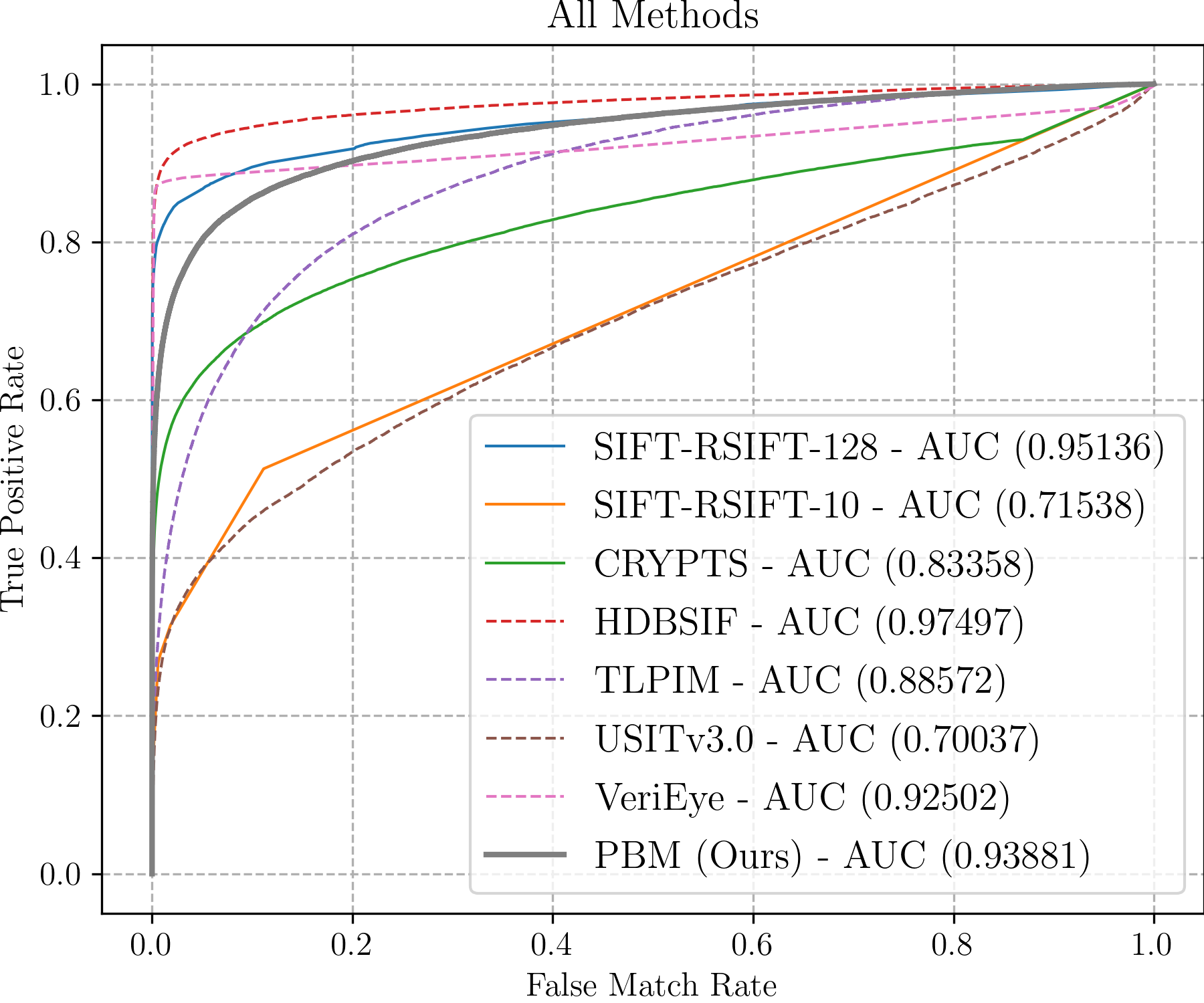}
    \end{center}
   \caption{Results for all baselines plus our proposed method (PBM). Dashed lines represent non-interpretable methods, solid lines represent human-interpretable methods, thicker solid line is our PBM method.  }
\label{fig:results}
\end{figure}

\begin{table}[!htb]
\centering
\caption{Equal Error Rates and decidability (d') scores. The proposed method compares favorably to the commercial (non-human-interpretable) VeriEye, and is best in terms of the d' score among all human-interpretable methods).}
\label{tab:dprime}
\begin{tabular}{c|c|c}
\textbf{Method}     & \textbf{EER (\%)}    & \textbf{d' score} \\ \hline\hline
SIFT-RSIFT-128 & 10.4 & 1.39              \\ \hline
SIFT-RSIFT-10 & 35.4 & 0.98              \\ \hline
Crypts \cite{7422104}       & 23.2 & 0.94              \\ \hline
HDBSIF \cite{Czajka_WACV2019}        & \textbf{6.4} & \textbf{2.54}     \\ \hline
TLPIM \cite{Kuehlkamp_WACVW_2022}        & 19.5 & 1.51              \\ \hline
USIT v3.0 \cite{rathgeb2016design}    & 35.8 & 0.82              \\ \hline
VeriEye \cite{Verieye}      & 11.1 & 1.29              \\ \hline\hline
\textbf{PBM (proposed)}     & \textbf{12.8} & \textbf{2.08}     \\
\end{tabular}
\end{table}

\subsection{Algorithms Compared To}

To compare the performance of the proposed method with state-of-the-art iris recognition, a set of baseline experiments were conducted with a variety of methods: human-interpretable and non-human-interpretable, deep learning and handcrafted, as well as commercial and open-source.

\vskip1mm\noindent
{\bf TLPIM} (Triplet Loss Postmortem Iris Model) \cite{Kuehlkamp_WACVW_2022} is a deep-learning based post-mortem iris matching approach that uses Class Activation Mapping to visualize important regions for post-mortem iris recognition. 
    
\vskip1mm\noindent
{\bf VeriEye} \cite{Verieye} is a popular commercial iris recognition tool produced by Neurotechnology. VeriEye uses Taylor expansion to extract image features that are then compared using a metric called ``elastic similarity'' in which impostor pairings produce results near zero.
    
\vskip1mm\noindent
{\bf USIT v3.0}  \cite{rathgeb2016design} is an open-source academic tool that implements Daugman-style iris recognition, using iris codes to calculate the Hamming distance between images. The configuration as suggested by the USIT authors was used: segmentation using CAHT, feature extraction from Ma \etal \cite{ma2004efficient} and matching using Hamming distance.
    
\vskip1mm\noindent
{\bf HDBSIF} (Human-Driven BSIF) \cite{Czajka_WACV2019} uses the ICA-trained filters, as in the original BSIF pipeline \cite{Kannala_ICPR_2012}, for extracting iris features. Two core differences with an original BSIF pipeline are (a) filters trained on iris image patches extracted from an eye-tracking device for people comparing iris samples, and (b) using binarized filter responses directly as iris codes, instead of comparing histograms of BSIF codes.
    
\vskip1mm\noindent
{\bf Keypoint-based.} 
    SIFT-RSIFT is a combination of general-purpose SIFT~\cite{lowe2004distinctive} and a more accurate variation of its 128-dimensional feature descriptor, a.k.a.~Root SIFT~\cite{arandjelovic2012three}. This solution leverages geometrically consistent content matching~\cite{moreira2018image} to find pairs of keypoints across two compared irises that present small feature-wise $L_2$ distances and high position equivalence (a.k.a.~matches). Inspired by fingerprint minutiae matching~\cite{galton1895fingerprint}, the number of matches is used to express the similarity between the two irises. Genuine pairs are expected to present large numbers of matches, while impostors are expected to present small numbers. To obtain robustness to the post-mortem collapse, keypoints are extracted from normalized irises. In the experiments, we explore two numbers of extracted keypoints, 128 (SIFT-RSIFT-128) and ten (SIFT-RSIFT-10).

\vskip1mm\noindent
{\bf Crypts} \cite{7422104} method implements detection and automatic matching of the iris crypts -- features that can be easily interpreted by humans. The designed matching scheme is able to handle potential topological changes in the detection of the same crypt in different images.

\subsection{Results}

From the ROC curves in Fig.~\ref{fig:results}, the best performing method on the held-out testing set ({\it dataset 2}) is HDBSIF. This is consistent with the results in
\cite{Boyd2020} for post-mortem iris recognition. However, HDBSIF is not human-interpretable, as it gives no indication of what features were important in the matching.  Instead, it uses the entire available iris surface for the match. Thus, it is not directly applicable to a forensic scenario in which human examiner's assessment is needed and the explanation as to why the pair of irises match, or not, is critically important. 

The best performing method that supplies justification of the decision is SIFT-RSIFT-128. This method returns pairs of matching keypoints and singular non-matching keypoints from the two sets of 128 keypoints. The AUC achieved by this method is $0.951$, narrowly outperforming our proposed approach that attains an AUC of $0.939$. However, because SIFT-RSIFT-128 uses 128 keypoints on each iris, the matching visualization may become cluttered and the interpretability is reduced due to the large number of extracted regions. 
Moreover, as in typical SIFT-like approaches, keypoints are represented by their central locations and compulsorily regular neighborhoods (usually circular or rectangular, thus ignoring the shapes of the compared regions during matching), making them not anatomy-driven and less human-interpretable.
The characteristics of being general-purpose and describing regular neighborhoods also reduce the robustness of keypoint-based approaches to the pupil dynamicity and post-mortem collapse of the irises (which are both visually non-linear and complex phenomena).
As a consequence, keypoint-based solutions must be applied over normalized irises, again hindering human-interpretability.
When the number of keypoints in the SIFT-RSIFT approach is reduced to be the same as for PBM, and thus its results are less cluttered, the performance decreases significantly to an AUC of $0.715$ (see SIFT-RSIFT-10).
Lastly, SIFT-RSIFT keypoints are not iris inspired and thus may not appear salient to a human examiner. 

In our proposed PBM method, the feature extractor is trained from human-annotated iris patches used in matching.
In addition to generating  more human-understandable features, this apparently brings a set of very discriminative features. Surprisingly, the PBM approach outperforms deep learning-based method ({\it TLPIM}), commercial ({\it VeriEye}) and Daugman-like approaches ({\it USITv3.0}), while also displaying interpretablity. The closest method in terms of interpretability to PBM is {\it Crypts}.  
However, as seen in Fig. \ref{fig:results} the performance of {\it Crypts} is significantly worse than PBM in the post-morterm iris recognition regime (note that the Crypts method was desined for live, not cadaver irises, hence its lower performance is understandable). Additionally, the top-performing baselines ({\it HDBSIF} \& {\it SIFT-RSIFT}) use iris normalization, which reduces interpretability as the iris texture is transformed to polar coordinate system. Our patch-based matching works with original images as if it were performed by a forensic examiner. 

Tab.~\ref{tab:dprime} shows the d-prime values for all methods. The d-prime metric measures the separability between the mean of the genuine comparisons and impostor comparisons, with a higher value indicating better separation and thus better performance. A higher d-prime also means more consistency across matches and more reliability. The best-performing method with regards to d-prime is again HDBSIF. The best-performing interpretable method and second best overall is the proposed Patch-Based Matching. This shows that although there is a sacrifice in performance compared to HDBSIF, the PBM approach adds human interpretability yet still performs reliably and predictably. 

\section{Conclusions}

This work introduces a new algorithm for post-mortem iris recognition, designed to (1) produce human-interpretable results and (2) achieve high accuracy in post-mortem iris matching.
Our foundation for producing human-interpretable results is a two-stage experimental data collection in which human examiners decide if a pair of iris images is from the same eye or not, and annotate image regions that support their decision. We find that the decisions of the verification stage are more accurate than those of the initial matching stage, with the improvement coming from more accurate classification of impostor pairs.
The decisions of the verification stage also have a much lower frequency of ``unsure'' results.
This implies that some large fraction of ``unsure'' judgements are examiner-dependent rather than a general result of available image features. 

This experiment produces important and useful conclusions on its own concerning how to achieve high accuracy in human evaluation of pairs of iris images.
And the iris image annotations collected in this experiment enable the training of the deep CNN to detect iris image features that are natural to human interpretation. Our proposed algorithm for automated matching of post-mortem iris images is evaluated against various state-of-the-art methods, some traditional, some designed for human interpretability, and some designed for post-mortem iris matching.
Comparing algorithms on a publicly-available dataset of post-mortem iris images, our proposed algorithm achieves the second-highest d-prime among the algorithms evaluated.
However, the algorithm with the highest d-prime uses a much larger number of features, and the features are not as directly human-interpretable.

Achieving a useful level of human interpretability almost always involves some tradeoff with accuracy.
In forensic post-mortem iris matching, human interpretability is essential.
Our proposed approach demonstrates minimal tradeoff with accuracy in the context of post-mortem iris recognition, while being designed from ground-up to display human-interpretable feature regions and matching. Source codes of the proposed method and trained models are being made available with this paper to contribute to the biometric community with human-interpretable, forensic-specific open-source iris recognition methods.

{\small
\bibliographystyle{ieee_fullname}
\bibliography{egbib}
}

\appendix

\section{Annotation Tool.}

\begin{figure*}
        \centering
        \includegraphics[width=0.9\textwidth]{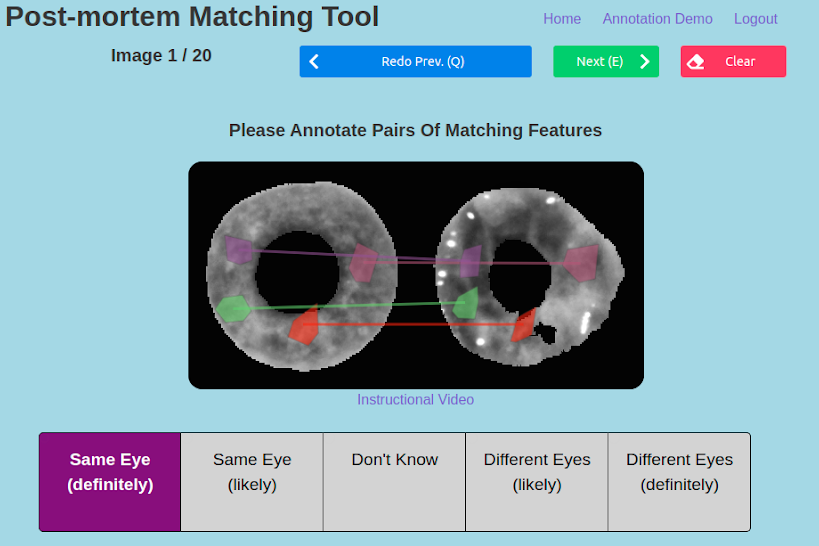}
        \caption{Screen shot of the Annotation Tool used to collect human decisions and annotations in Step 1 (``Match Evaluation'').}
        \label{fig:annot_step1}
\end{figure*}

\begin{figure*}
        \centering
        \includegraphics[width=0.9\textwidth]{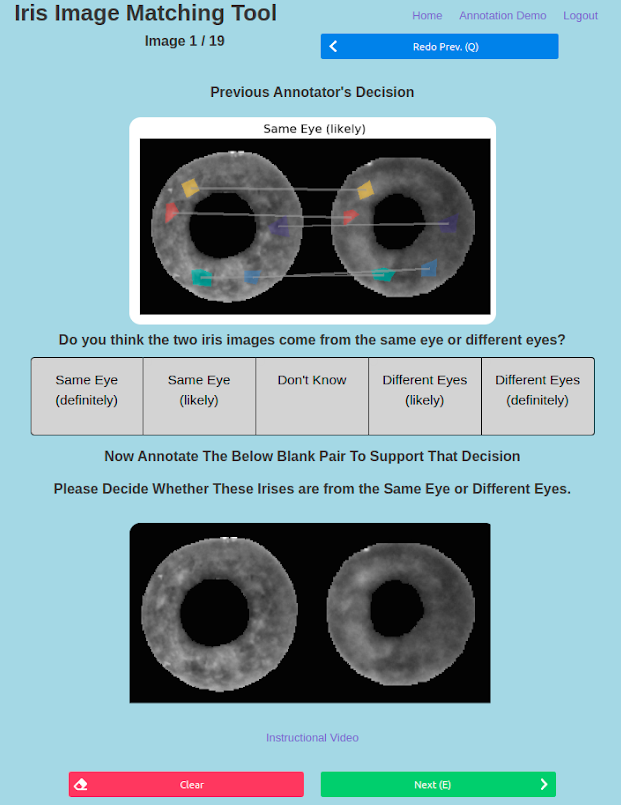}
        \caption{Screen shot of the Annotation Tool used to collect human decisions and annotations in Step 2 (``Match Verification'').}
        \label{fig:annot_step2}
\end{figure*}

Two versions of the Annotation Tool were created for the purpose of this work. The first version, shown in Fig. \ref{fig:annot_step1} and used in Step 1 (``Match Evaluation''), collects annotations and decision {\bf without} the knowledge of the outcomes from the previous examiner. The second version, shown in Fig. \ref{fig:annot_step2} and in Step 2 (``Match Verification''), collects annotations and decision {\bf with} the outcomes (annotations and decision) from the previous examiner available to the next subject.

\section{Additional Annotation Collection Results}

\begin{table}[!htb]
\caption{Numbers of decisions that have been changed in Step 2 by subjects seeing outcomes from the previous examiners.}
\label{tab:decision_changes}
\begin{center}
\begin{tabular}{l|c|c}
& \textbf{Incorrect } & \textbf{Correct } \\
& \textbf{to Correct} & \textbf{to Incorrect} \\\hline\hline
\textbf{Genuine to Impostor} & 153                           & 46                            \\ \hline
\textbf{Impostor to Genuine} & 92                            & 109                           \\ \hline\hline
\textbf{Unsure to Genuine}   & 27                            & \cellcolor[HTML]{cccccc}      \\ \hline
\textbf{Unsure to Impostor}  & 63                            & \cellcolor[HTML]{cccccc}      \\ \hline\hline
\textbf{Genuine to Unsure}   & \cellcolor[HTML]{cccccc}      & 15                            \\ \hline
\textbf{Impostor to Unsure}  & \cellcolor[HTML]{cccccc}      & 15                            \\ \hline
\end{tabular}
\end{center}
\end{table}

As illustrated in Table~\ref{tab:decision_changes}, more annotators changed their responses from an incorrect decision to a correct decision in the ``Match Verification'' trials (Step 2) than from correct to incorrect. 
In addition, three times as many Step 2 annotators changed the previous annotator's decision from unknown to a correct decision, than from a correct decision to unknown. 

\section{MaskRCNN training configuration}

We followed the implementation of MaskRCNN found at \url{https://github.com/matterport/Mask_RCNN}. The specific values of hyperparameters used in this work are listed in Table \ref{tab:maskrcnn}.

\begin{table*}[!htb]
\caption{Values of the hyperparameters of the Mask RCNN model used in this work.}\vskip2mm
\begin{center}
\label{tab:maskrcnn}
\begin{tabular}{l|l}
\hline
\textbf{BACKBONE}                        & resnet50                                                                                                                           \\ \hline
\textbf{BACKBONE\_STRIDES}               & {[}4, 8, 16, 32, 64{]}                                                                                                             \\ \hline
\textbf{BATCH\_SIZE}                     & 10                                                                                                                                 \\ \hline
\textbf{BBOX\_STD\_DEV}                  & {[}0.1 0.1 0.2 0.2{]}                                                                                                              \\ \hline
\textbf{COMPUTE\_BACKBONE\_SHAPE}        & None                                                                                                                               \\ \hline
\textbf{DETECTION\_MAX\_INSTANCES}       & 100                                                                                                                                \\ \hline
\textbf{DETECTION\_MIN\_CONFIDENCE}      & 0                                                                                                                                  \\ \hline
\textbf{DETECTION\_NMS\_THRESHOLD}       & 0.3                                                                                                                                \\ \hline
\textbf{FPN\_CLASSIF\_FC\_LAYERS\_SIZE}  & 1024                                                                                                                               \\ \hline
\textbf{GPU\_COUNT}                      & 1                                                                                                                                  \\ \hline
\textbf{GRADIENT\_CLIP\_NORM}            & 5.0                                                                                                                                \\ \hline
\textbf{IMAGES\_PER\_GPU}                & 10                                                                                                                                 \\ \hline
\textbf{IMAGE\_CHANNEL\_COUNT}           & 3                                                                                                                                  \\ \hline
\textbf{IMAGE\_MAX\_DIM}                 & 256                                                                                                                                \\ \hline
\textbf{IMAGE\_META\_SIZE}               & 14                                                                                                                                 \\ \hline
\textbf{IMAGE\_MIN\_DIM}                 & 256                                                                                                                                \\ \hline
\textbf{IMAGE\_MIN\_SCALE}               & 0                                                                                                                                  \\ \hline
\textbf{IMAGE\_RESIZE\_MODE}             & square                                                                                                                             \\ \hline
\textbf{IMAGE\_SHAPE}                    & {[}256 256 3{]}                                                                                                                    \\ \hline
\textbf{LEARNING\_MOMENTUM}              & 0.9                                                                                                                                \\ \hline
\textbf{LEARNING\_RATE}                  & 0.001                                                                                                                              \\ \hline
\textbf{LOSS\_WEIGHTS}                   & \{'rpn\_class\_loss': 1.0, 'rpn\_bbox\_loss': 1.0, 'mrcnn\_class\_loss': 1.0,\\ & 'mrcnn\_bbox\_loss': 1.0, 'mrcnn\_mask\_loss': 3.0\} \\ \hline
\textbf{MASK\_POOL\_SIZE}                & 14                                                                                                                                 \\ \hline
\textbf{MASK\_SHAPE}                     & {[}28, 28{]}                                                                                                                       \\ \hline
\textbf{MAX\_GT\_INSTANCES}              & 30                                                                                                                                 \\ \hline
\textbf{MEAN\_PIXEL}                     & {[}50. 50. 50.{]}                                                                                                                  \\ \hline
\textbf{MINI\_MASK\_SHAPE}               & (56, 56)                                                                                                                           \\ \hline
\textbf{NAME}                            & iris\_feature\_finetuned                                                                                                           \\ \hline
\textbf{NUM\_CLASSES}                    & 2                                                                                                                                  \\ \hline
\textbf{POOL\_SIZE}                      & 7                                                                                                                                  \\ \hline
\textbf{POST\_NMS\_ROIS\_INFERENCE}      & 1000                                                                                                                               \\ \hline
\textbf{POST\_NMS\_ROIS\_TRAINING}       & 2000                                                                                                                               \\ \hline
\textbf{PRE\_NMS\_LIMIT}                 & 6000                                                                                                                               \\ \hline
\textbf{ROI\_POSITIVE\_RATIO}            & 0.33                                                                                                                               \\ \hline
\textbf{RPN\_ANCHOR\_RATIOS}             & {[}0.5, 1, 2{]}                                                                                                                    \\ \hline
\textbf{RPN\_ANCHOR\_SCALES}             & (8, 16, 32, 64, 128)                                                                                                               \\ \hline
\textbf{RPN\_ANCHOR\_STRIDE}             & 1                                                                                                                                  \\ \hline
\textbf{RPN\_BBOX\_STD\_DEV}             & {[}0.1 0.1 0.2 0.2{]}                                                                                                              \\ \hline
\textbf{RPN\_NMS\_THRESHOLD}             & 0.9                                                                                                                                \\ \hline
\textbf{RPN\_TRAIN\_ANCHORS\_PER\_IMAGE} & 256                                                                                                                                \\ \hline
\textbf{STEPS\_PER\_EPOCH}               & 200                                                                                                                                \\ \hline
\textbf{TOP\_DOWN\_PYRAMID\_SIZE}        & 256                                                                                                                                \\ \hline
\textbf{TRAIN\_BN}                       & False                                                                                                                              \\ \hline
\textbf{TRAIN\_ROIS\_PER\_IMAGE}         & 256                                                                                                                                \\ \hline
\textbf{USE\_MINI\_MASK}                 & False                                                                                                                              \\ \hline
\textbf{USE\_RPN\_ROIS}                  & True                                                                                                                               \\ \hline
\textbf{VALIDATION\_STEPS}               & 100                                                                                                                                \\ \hline
\textbf{WEIGHT\_DECAY}                   & 0.01                                                                                                                               \\ \hline
\end{tabular}
\end{center}
\end{table*}

\end{document}